\documentclass{article}
 
\usepackage{arxiv}
 
\usepackage[utf8]{inputenc} % allow utf-8 input
\usepackage[T1]{fontenc}    % use 8-bit T1 fonts
\usepackage{hyperref}       % hyperlinks
\usepackage{url}            % simple URL typesetting
\usepackage{booktabs}       % professional-quality tables
\usepackage{amsfonts}       % blackboard math symbols
\usepackage{nicefrac}       % compact symbols for 1/2, etc.
\usepackage{microtype}      % microtypography
\usepackage{graphicx}
\usepackage[numbers,sort&compress]{natbib}
\usepackage{doi}
 
\usepackage{amsmath,amssymb}
\usepackage{amsthm}
\usepackage{mathrsfs}
\usepackage{xcolor}
\usepackage{subfig}
\usepackage{textcomp}
\usepackage{multirow}
\usepackage{algorithm}
\usepackage{algorithmicx}
\usepackage{algpseudocode}
\usepackage{placeins}

\newcommand{\tblwidth}{\linewidth}
 
\theoremstyle{plain}

\theoremstyle{definition}

\title{Early Language Learning via Spreading Activation and Category Exploration in Complex Networks}
 
\renewcommand{\shorttitle}{Early Language Learning via Spreading Activation and Category Exploration}
\renewcommand{\headeright}{Citraro}
 
\author{
	\href{https://orcid.org/0000-0002-5021-4790}{\includegraphics[scale=0.06]{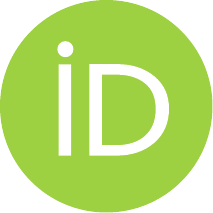}\hspace{1mm}Salvatore Citraro} \\
	Institute of Information Science and Technologies ``A. Faedo'' (ISTI)\\
	National Research Council (CNR)\\
	Via G. Moruzzi, 1, 56124 Pisa, Italy \\
	\texttt{salvatore.citraro@isti.cnr.it}
}
 
\hypersetup{
pdftitle={Early Language Learning via Spreading Activation and Category Exploration in Complex Networks},
pdfsubject={q-bio.NC, cs.CL, physics.soc-ph},
pdfauthor={Salvatore Citraro},
pdfkeywords={complex networks, network diffusion, spreading activation, early language learning, vocabulary development},
}
 
\begin{document}
\maketitle

\begin{abstract}
Is word acquisition in children uneven with respect to semantic and lexical categories?
To answer this question, we model early language learning as a search on a graph-based mental lexicon, driven by two interacting processes: spreading activation and an enforced exploration (rather than exploitation) of lexical categories.
We evaluate model performance on four languages (German, English, Dutch, and Rioplatense Spanish), using CDIs as ground-truth data for lexical categories, normative ages derived from the Wordbank repository, and state-of-the-art resources for reconstructing graphs of word similarities.
We find that spreading activation outperforms a shortest path baseline in simulating normative word acquisition.
At the category level, we highlight complex transitions between CDIs.
By studying their sequences in terms of burstiness and average persistence time within the same CDI, we find that spreading activation better captures the exploration dynamics observed empirically.
Overall, our findings suggest that vocabulary development can be understood through the non-trivial interplay between activation dynamics and some degree of constraints regulating the ``visiting'' of lexical categories in complex networks.
\end{abstract}

% keywords can be removed

%%--------------------------------------------------------------------
\section{Introduction}\label{sec:intro}
%%--------------------------------------------------------------------

Early language learning includes the process by which toddlers become familiar with words and their use in different contexts \cite{wittgenstein1953, firth1957synopsis, weizman2001lexical, bloom2002, lenci2022comparative}.
Evidence suggests that vocabulary development exhibits patterns, with words and categories emerging at different stages across children and languages \cite{fenson2007macarthur, storkel2009developmental, braginsky2019consistency}.
However, detecting and explaining word acquisition regularities remains an open problem in cognitive science and computational linguistics.
Why certain words and their associated categories are learned earlier or later than others is the central focus of this work, which we address through the lens of complex network science \cite{stella2017multiplex, siew2019cognitive, hills2022mind, haim2026cognitive}.

In recent decades, the mental lexicon (the cognitive repository of human word knowledge) has been represented as a complex system, more specifically as a complex network, where words are the nodes of a graph and the linguistic relations between words are the edges \cite{steyvers2005large, beckage2016language, stella2017multiplex, siew2019cognitive, haim2026cognitive}.
Representing language as a graph provides a computational realization of structural linguistics, in which linguistic units derive their meaning from their position within a system of relations \cite{de2017course}, thus extending our perspective on language by means of the properties of complex systems.
Across different network representations of language that encode phonological, syntactic, and semantic knowledge \cite{stella2024cognitive}, several of these properties emerge, from small-world organization to heavy-tailed degree distributions \cite{steyvers2005large, beckage2016language, de2019small, siew2019cognitive}.
These features can identify core words as network hubs and reveal a clustered organization of dense groups, such as graph cliques \cite{chern2025evidence, vitevitch2016phonological, stella2024cognitive}, and kernels, such as lexical viable clusters in multiplex enrichment \cite{stella2017multiplex}.

Furthermore, network representations of the mental lexicon have been useful in studying cognitive phenomena such as lexical retrieval, semantic fluency, and early language learning \cite{vitevitch2008can, hills2009longitudinal, hills2010associative, stella2017multiplex, siew2019cognitive}, suggesting that a complex structure can shape the dynamics of cognitive search, e.g., affecting which words are reached earlier or later during network navigation.
For instance, more central, better connected words require fewer steps to be accessed \cite{hills2009longitudinal, stella2017multiplex, siew2019cognitive}.
At the same time, the dynamics of network navigation need not reduce to simple heuristics.
Word retrieval in fluency tasks can be further modeled as a mix of exploitation and exploration of semantic ``patches'' \cite{hills2012optimal, hills2015foraging, hills2022mind}, analogous to animal foraging behavior while searching for food.
Similarly, spreading activation theory suggests that cognitive activation propagates from a set of seed nodes to their adjacent neighbors, weakening as network distance increases \cite{collins1975spreading}.
Spread activation can explain the cognitive aspects underlying false memory, semantic priming \cite{siew2019cognitive}, and it can quantify the interplay between phonological and semantic layers in mixed aphasic errors (e.g., saying \textit{rat} for \textit{cat}) within multiplex frameworks \cite{citraro2026spreadpy}.

When simulating early language learning with complex networks, much of the literature has focused on theoretical growth models \cite{beckage2019network}, with works relying on preferential attachment \cite{steyvers2005large} or on mixed mechanisms in which a word becomes more likely to be acquired as more of its neighbors are already known \cite{hills2009longitudinal, hills2010associative}.
Some models distinguish between preferential attachment underlying paradigmatic relations (e.g., \textit{cherry}-\textit{apple}) and random mechanisms shaping syntagmatic ones (e.g., \textit{\{cherry,apple\}}-\textit{red}) \cite{utsumi2015complex}.
The role of a random component is important when comparing the network structure of typical and late talkers, as also ``noise" can shape the small-world organization observed in the former but not in the latter \cite{beckage2011small}.
Vocabulary development has been investigated as a search on empirical lexicons as well, e.g., via random walks biasing the exploration towards ``empty'' semantic categories \cite{ferrer2018origins, citraro2023feature}.
These studies are motivated by work suggesting that toddlers do not acquire word classes uniformly \cite{fenson2007macarthur, braginsky2019consistency}, while others, however, claim that random walks could not provide a sufficient explanation for lexical acquisition \cite{beckage2012route}.

Building on all these findings, we propose here a computational model of early language learning that integrates a network process based on spreading activation (rather than random walks) \cite{collins1975spreading, siew2019spreadr, citraro2026spreadpy}, coupled with a mechanism of exploration that penalizes repeated selection of lexical categories already explored \cite{citraro2023feature}.
%These two mechanisms may appear opposed, as assuming that adjacent neighbors clustered around an activated word are more likely to become activated than distant ones, whereas category exploration can move the activation toward different network regions.
%Combining them is motivated by theoretical proposals suggesting that lexical acquisition may depend not only on word similarities, explicitly encoded by complex networks, but also on latent factors (e.g., inhibitory semantic priming effects \cite{arias2013s}, skyhook metaphors \cite{hills2022mind}, or higher-order paths \cite{rubinosimilarity}, among others) that have not yet been incorporated into computational models of early language learning.
Our algorithm is tested on four languages (German, English, Dutch, and Rioplatense Spanish), using complex cognitive networks of free associations, WordNet relations, and phonological similarities, and is evaluated against normative age of acquisition sequences and lexical categories extracted from Wordbank data, built upon the MacArthur-Bates Communicative Development Inventories \cite{fenson2007macarthur}.
This cross-linguistic design allows us to assess the robustness of our experiments beyond English, where most previous empirical work has been focused so far \cite{stella2017multiplex, citraro2023feature}.
%Across word- and category-level analyses, we find that predictions of normative age sequences are consistent across languages in some aspects and differ in others.

We find that the empirical sequences are better matched by spreading activation dynamics with high retention levels, namely a regime in which activation remains distributed across previously activated nodes rather than being wholly transferred to adjacent neighbors at each step.
Moreover, complex transitions emerge at the scale of lexical categories.
We analyze category transitions in terms of persistence and burstiness, which quantify, respectively, the average time spent exploiting the same category before switching, and the distribution of inter-times between occurrences of the same category \cite{goh2008burstiness}.
Under specific configurations, our model can reproduce transitions statistically significantly closer to the empirical ones than those generated by baselines.

The remainder of the work is organized as follows.
Section~\ref{sec:methods} describes in detail network data, lexical categories, age of acquisition ground-truth, and our algorithm.
Section~\ref{sec:res} presents the results of our experiments on the four languages, evaluating the goodness of our algorithm in recovering the empirical acquisition sequences and the exploration dynamics at the level of lexical categories.
Section~\ref{sec:disc} discusses implications, limitations, and future work.

%%--------------------------------------------------------------------
\section{Methods}\label{sec:methods}
%%--------------------------------------------------------------------

%\FloatBarrier
\subsection{Network Data}\label{subsec:net_data}

We represent a mental lexicon of word similarities as a complex network with aggregated information, including free associations, synonyms, antonyms, hypo/hypernyms, and phonemic similarities \cite{stella2024cognitive}.
These levels constitute the most common types of word similarity encoded in cognitive network tasks \cite{stella2017multiplex, siew2019cognitive, stella2024cognitive}, jointly capturing phonological competence (proxied by phonemic distances), and semantic knowledge (represented by all other relations mentioned above).

Specifically, for free associations, we collect data from the Small World of Words project (henceforth, SWoW)\footnote{\url{https://smallworldofwords.org/en/project/research}}.
We focus on the following releases for German, English, Dutch, and Rioplatense Spanish: SWOW-DE25 \cite{aeschbach2026small}, SWOW-EN18 \cite{de2019small}, SWOW-NL13 \cite{de2013better}, and SWOW-RP22 \cite{cabana2024small}.
Word association data are well-established models of semantic memory often represented as complex networks \cite{de2013better, stella2017multiplex}, namely as graphs $G=(N,E)$ where words are nodes ($N$) and free associations are edges ($E$).
Grounding our model of vocabulary development as a lexical retrieval process on a mental lexicon/complex network, the SWoW datasets provide a state-of-the-art relational structure.

Nevertheless, it is still possible to enrich it with other relations commonly recognized in linguistics and cognitive network science \cite{siew2019cognitive}.
Therefore, we integrate data from WordNet \cite{bond2013linking}: NLTK WordNet\footnote{\url{https://www.nltk.org/howto/wordnet.html}} for English, Dutch, and Spanish, and Open German WordNet \cite{siegel2021odenet}.
In particular, we include synonymy, antonymy, and hypo/hypernym relations, where synonymy captures overlap in word meaning, as in \textit{car-automobile}, antonymy encodes oppositional messages, as in \textit{hot-cold}, and hypo/hypernyms represent hierarchical relations, where a more specific concept (hyponym) is related to a more general one (hypernym), as in \textit{dog-animal}.
These levels add density to a network-based mental lexicon by encoding taxonomic organization and purely lexical relations \cite{stella2017multiplex}.
For simplicity, when a link appears across multiple levels, it is counted only once; that is, we consider an unweighted network without edge weights.

Finally, we incorporate phonological similarity by linking words having similar phonemic transcriptions.
Phonemes are obtained via the \textit{CMUdict}\footnote{\url{https://pypi.org/project/cmudict/}} library for English and the \textit{epitran}\footnote{\url{https://pypi.org/project/epitran/}} python package for the other languages.
Edges are added between word pairs whose edit distance $d$ between phoneme strings is below a fixed threshold.
We use $d \leq 2$, yielding a coarse-grained phonological layer that can capture broader perceptual similarity in toddlers.

\begin{table}[t!]
\centering
\begin{tabular*}{\tblwidth}{@{}lcc@{}}
\toprule
Language & $|N|$ & $|E|$ \\
\midrule
German ($de$)               & 432 & 2812 \\
English ($en$)              & 599 & 8245 \\
Dutch ($nl$)                & 446 & 4432 \\
Riopltns. Spanish ($sp$)    & 507 & 4441 \\
\bottomrule
\end{tabular*}
\caption{Number of words and edges of the lexical networks reconstructed.}
\label{tab:network_stats}
\end{table}

In total, we retrieve the number of nodes and edges summarized in Table~\ref{tab:network_stats}.
The size of each network, i.e., the vocabulary size $|N|$, is determined by the intersection between the ground-truth vocabulary (cf., Section~\ref{subsec:cdi}) and the set of words covered by the corresponding SWoW dataset.

%\FloatBarrier
\subsection{Lexical Categories and Age of Acquisition}\label{subsec:cdi}

Vocabularies are obtained from the Stanford Wordbank\footnote{\url{https://wordbank.stanford.edu/data/?name=item_data}}, an open database of toddlers' language development, built upon the MacArthur-Bates Communicative Development Inventories (henceforth, CDIs) \cite{fenson2007macarthur}.
CDIs are averaged parent reports describing children's developing abilities in early language learning, including vocabulary development.
Wordbank spans dozens of languages, but the language overlap with SWoW currently restricts our analysis to the four languages already mentioned above: German, English, Dutch, and Rioplatense Spanish.
For each word, we retrieve two types of ground-truth information: (i) its lexical category as defined by the CDI form (e.g., \textit{animals}, \textit{toys}, \textit{pronouns}) and (ii) its average age of acquisition.

Lexical categories are largely consistent across the four languages.
However, as said in the previous paragraph, the size of each vocabulary, $|N|$, is determined by the overlap with Wordbank and SWoW.
This could result in some CDIs being underrepresented.
To mitigate this issue, we exclude CDIs appearing at most 4 times; exclusion in one language does not imply exclusion in others.
This filtering step should improve the consistency and significance of our experiments.

Ages range from 16 to 36 months for each individual word produced by each child.
However, we average age at the word level across children.
Then, from these averages we derive the acquisition rank of each word, namely a ground-truth sequence $R_{gt}$ that our model aims to predict for each language: $R_{{gt}_{de}}$, $R_{{gt}_{en}}$, $R_{{gt}_{nl}}$, and $R_{{gt}_{sp}}$.
The procedure for building each $R_{gt}$ is simple: Each word is assigned its empirical average age of acquisition and then sorted accordingly.
For example, in English, \textit{mommy} and \textit{daddy} have the lowest average values, 23.068 and 23.075 months, respectively, whereas \textit{yesterday} and \textit{yourself} have the highest ones, 27.176 and 27.189 months.
Consequently, the first two elements of the rank $R_{{gt}_{en}}$ are \textit{mommy} and \textit{daddy}, while the last two are \textit{yesterday} and \textit{yourself}.

%\FloatBarrier
\subsection{Vocabulary Development Algorithm}\label{subsec:algo}

We model vocabulary development as a selection process on a network-based mental lexicon $G=(N,E)$, where nodes $v \in N$ represent words and edges $e=\{u,v\} \in E$ represent relations between pairs of words.
Our aim is to build a vocabulary sequence $R_{pred}$, able to match the previously described ground-truth sequence $R_{gt}$.
At each iteration $t \in \{1, \ldots ,|N|\}$, the next word to be acquired is selected by maximizing the following score:

\begin{equation}
\label{eq:linear_comb}
    \gamma_v^{(t)} = (1 - \mu)\,\alpha_v^{(t)} + \mu\,\delta_v^{(t)},
\end{equation}

with $\alpha_v^{(t)}$ representing the node's structural score at iteration $t$, $\delta_v^{(t)}$ denoting the component accounting for lexical exploration, and $\mu \in [0,1]$ controlling the trade-off between structural- and lexical-based selection importance.
When $\mu = 0$, the process is driven only by the structural component; when $\mu = 1$, only by lexical exploration.

Hence, at iteration $t$, the word selected for acquisition is:

\begin{equation}
v_t^* =
\underset{v \,\in\, N \setminus N^{(t-1)}}{\arg\max}
\; \gamma_v^{(t)},
\label{eq:node_sel}
\end{equation}

where $N^{(t-1)}$ is the set of words contained in the ordered sequence $R_{\text{pred}}^{(t-1)}$, i.e., the set of already acquired words.
The selected word $v_t^*$ is then inserted into the predicted acquisition ordering $R_{pred}$, and the process iterates until $|R_{\text{pred}}^{(t)}| = |N|$, i.e., until all words have been acquired.

A pseudocode of the algorithm, reporting spreading activation as the structural score, is presented in Algorithm~\ref{alg:1}.
The reader may find it useful to consult it after completing this section.

%\FloatBarrier
\subsubsection{Structural Dynamics ${\alpha}_v$}

\paragraph{Spreading Activation}

We follow the spreading activation theory \cite{collins1975spreading}, according to which cognitive activation spreads from a seed node to its adjacent neighbors, and iteratively propagates through the network over a finite number of discrete time steps, say $t_{s_{max}}$.
Note that we distinguish here between iteration $t \in \{1, \ldots ,|N|\}$ and iteration $t_s \in \{1, \ldots ,t_{s_{max}}\}$, where the former is the time of the vocabulary growth (see above) and the latter is the inner time of a spreading activation run.
Several computational models have been proposed in the literature for simulating spreading activation mechanisms \cite{siew2019spreadr, citraro2026spreadpy}.
Relying on them, we use the following formula for modeling lexical energy diffusion from an activated word to its neighbors:

\begin{equation}
\varphi(u, v) = \frac{\epsilon_{u,t_{s}} \cdot (1 - r)}{\left| \Gamma_u \right|},
\end{equation}

which refers to the energy $\epsilon_{u,t_{s}}$ transferred from node $u$ to neighbor $v \in \Gamma_u$ at time $t_s$, being $\Gamma_u$ the set of $u$'s neighbors, and $r$ a retention parameter keeping a portion of energy within $u$.
Thus, a quantity of the activation energy is distributed uniformly across adjacent neighbors, while the remaining fraction $r$ remains at the source node.
The diffusion process is initialized by assigning a fixed activation energy $\epsilon_0$ to the current seed node, i.e., the most recently acquired word, while all other nodes start with zero activation.
Then, for each candidate node $v$, we consider its peak activation over spreading time, i.e., the maximum activation value reached by the node during the current spreading activation run after completed $t_{s_{max}}$ iterations, as follows:

\begin{equation}
    {\alpha}_v^{(t)} = \max_{t_{s}} \, \epsilon_{v,t_{s}}.
    \label{eq:struct_score}
\end{equation}

Note that ${\alpha}_v^{(t)}$ can be normalized prior to combination with $\delta_v$ (see later), to ensure comparability of terms in Eq. \ref{eq:linear_comb}, e.g., with a min-max normalization.

Consider starting from the word \textit{mom} as a seed node triggering a first run of the spreading activation process.
When $\mu=0$, at the end of diffusion, the word that reaches the highest ${\alpha}_v$, i.e., the word with the highest activation $\epsilon$ in the shortest time $t_{s}$, say \textit{dad}, is selected as the next predicted word and added to the vocabulary growth sequence.

\paragraph{Shortest Paths}

As a baseline to compare against spreading activation, we consider an alternative structural mechanism based on shortest paths, inspired by the role of closeness centrality in detecting structural patterns of early language learning \cite{stella2017multiplex}.

Given the current seed node $u$, the structural score of a candidate node $v$ is defined as follows:

\begin{equation}
    {\alpha}_v^{(t)} = \frac{1}{d(u, v)},
    \label{eq:struct_score_baseline}
\end{equation}

where $d(u, v)$ denotes the shortest path length between $u$ and $v$ in the network. Note that $d(u, v)=0$ only when $u=v$, a case that can not occur in our formulation.
Also in this case, note that ${\alpha}_v^{(t)}$ can be normalized prior to combination with $\delta_v$ (see later), to ensure comparability of terms in Eq. \ref{eq:linear_comb}, e.g., with a min-max normalization.
Finally, note that, unlike spreading activation, this score depends only on the static graph structure, thus it does not induce a temporal activation trajectory.

%\FloatBarrier
\subsubsection{Semantic Dynamics $\delta_v$}

Now consider relying not only on the structural component ${\alpha}_v^{(t)}$ but also taking into account persistence within the same lexical category.
Motivated by previous findings suggesting an interplay between the exploration and exploitation of CDI categories \cite{citraro2023feature}, favoring the former, we enforce here a mechanism that biases selection toward underrepresented lexical categories.
In other words, a highly activated or central word may be preferred over another highly activated or central word if the CDI of the former has been observed less frequently than the CDI of the latter.
In this exploration, starting from the seed node \textit{mom}, the algorithm could prefer \textit{dog} over \textit{dad} because the former belongs to a category (animals) that has been observed less than the latter (people, to which both \textit{mom} and \textit{dad} belong).

Let $\mathcal{C} = \{c_1, c_2, \ldots, c_K\}$ denote the set of $K$ CDI lexical categories, and let $f_{c_k}^{(t)}$ denote the number of times category $c_k \in \mathcal{C}$ has appeared in the predicted ordering up to time $t$.
Categories are ranked in ascending order of $f_{c_k}^{(t)}$, so that less frequent categories receive higher priority.
For a node $v$ belonging to category $c_k$, the lexical exploration score is defined as:

\begin{equation}
    \delta_v^{(t)} = 1 - \frac{\text{rank}(f_{c_k}^{(t)})}{K - 1},
    \label{eq:exp_score}
\end{equation}

where $rank(\cdot) \in \{0, \ldots, K-1\}$ is computed over $f_{c_k}^{(t)}$ in ascending order so that $\delta_v^{(t)} = 1$ for nodes in the least represented category and $\delta_v^{(t)} = 0$ for nodes in the most represented one.

In case of ties in $\delta_v^{(t)}$, which may occur in the early stages of the process when most categories are equally unobserved (e.g., $f_{c_k}^{(t)} = 0$ for multiple categories), the rank is assigned randomly among the tied slots: if $m$ categories share the same frequency, they are randomly permuted across $m$ consecutive rank positions.

\begin{algorithm}[t]
\caption{Vocabulary Development Algorithm with Spreading Activation}
\label{alg:1}
\begin{algorithmic}[1]
\Require Graph $G=(N,E)$, initial seed $v_0$, parameters $r$, $\mu$, $t_{s_{max}}$
\ \\ \
\State \textbf{Initialization ($t=1$):}
\State \hspace{\algorithmicindent} $R_{\text{pred}} \leftarrow [v_0]$
\State \hspace{\algorithmicindent} Initialize category counts $f_c \leftarrow 0 \;\forall c$
\State \hspace{\algorithmicindent} $f_{c(v_0)} \leftarrow f_{c(v_0)} + 1$
\ \\ \
\For{$t = 2$ to $|N|$} \Comment{vocabulary growth time $t$}
    \State $u \leftarrow \mathrm{last}(R_{\text{pred}})$
    \State Run spreading activation from current seed $u$ for $t_{s_{max}}$ steps\Comment{spreading activation time $t_s$}
    \State Compute structural scores $\alpha_v^{(t)}$\Comment{from Eq. \ref{eq:struct_score}}
    \State Compute exploration scores $\delta_v^{(t)}$ \Comment{from Eq. \ref{eq:exp_score}}

    \State $\gamma_v^{(t)} \leftarrow (1-\mu)\alpha_v^{(t)} + \mu \delta_v^{(t)}$\Comment{Eq. \ref{eq:linear_comb}}

    \State $v_t^* \leftarrow \arg\max_{v \in  N \setminus N^{(t-1)}} \gamma_v^{(t)}$\Comment{Eq. \ref{eq:node_sel}}

    \State Append $v_t^*$ to $R_{\text{pred}}$
    \State $f_{c(v_t^*)} \leftarrow f_{c(v_t^*)} + 1$

\EndFor
\\ \ \\
\State \Return $R_{\text{pred}}$

\end{algorithmic}
\end{algorithm}

%%--------------------------------------------------------------------
\section{Results}\label{sec:res}
%%--------------------------------------------------------------------

%\FloatBarrier
\subsection{Modeling Setups}
\label{subsec:setup}
\smallskip
\noindent \textbf{Spreading activation}.
For instances of the algorithm with $\alpha_v^{(t)}$ driven by spreading activation, we use the following parameters: retention $r$ in the range $[0, 0.9]$ with bins of $0.1$, maximum diffusion time of a single spreading run $t_{s_{max}}=50$, and $\mu$ in the range $[0, 1]$ with bins of $0.1$.
Moreover, in all diffusion runs, the current seed node at each iteration is assigned a fixed initial activation $\epsilon_0 = 1.0$.
Note that an initial activation is required at the beginning of each run to trigger the spreading process.
Also importantly, only the very first iteration requires an ``externally'' specified seed node $v_0 \in R_{gt}$.
Therefore, we use the first word of the empirical acquisition sequence, namely \textit{mama} for German, \textit{mommy} for English, \textit{die} for Dutch, and \textit{no} for Spanish.
This is the only step in which information from the ground-truth ordering is injected; all other steps are generated by the model according to Eq.~\ref{eq:linear_comb}.
Let us explain this clearly.
Suppose that, for English, the first diffusion run is initialized with the externally provided seed \textit{mommy}. After the first iteration of the spreading process, the model selects \textit{yourself} as the next word to be acquired.
At this point, the second diffusion run is initialized using \textit{yourself} as the new seed node, and not \textit{daddy}, which is the second empirical word.
Therefore, from this step onward, the evolution of the acquisition sequence depends entirely on the model's own previous predictions.

\smallskip

\noindent \textbf{Baselines}.
For the algorithm with $\alpha_v^{(t)}$ driven by shortest paths, since it is parameter-free and dependent on the algorithmic time alone, we vary only $\mu$ in the range $[0, 1]$ with bins of $0.1$.
To ensure consistency with the spreading logic, here we also consider the first word in the ground-truth sequence as the initial seed node.

Finally, for testing both against random word acquisition, we build a null model in which the empirical frequencies of CDI are preserved but the temporal ordering of words is randomized.
For each language, we generate $n=2000$ random permutations of the ground-truth sequence; the reshuffling only reorders words, without altering their lexical categories.

\begin{figure}[t!]
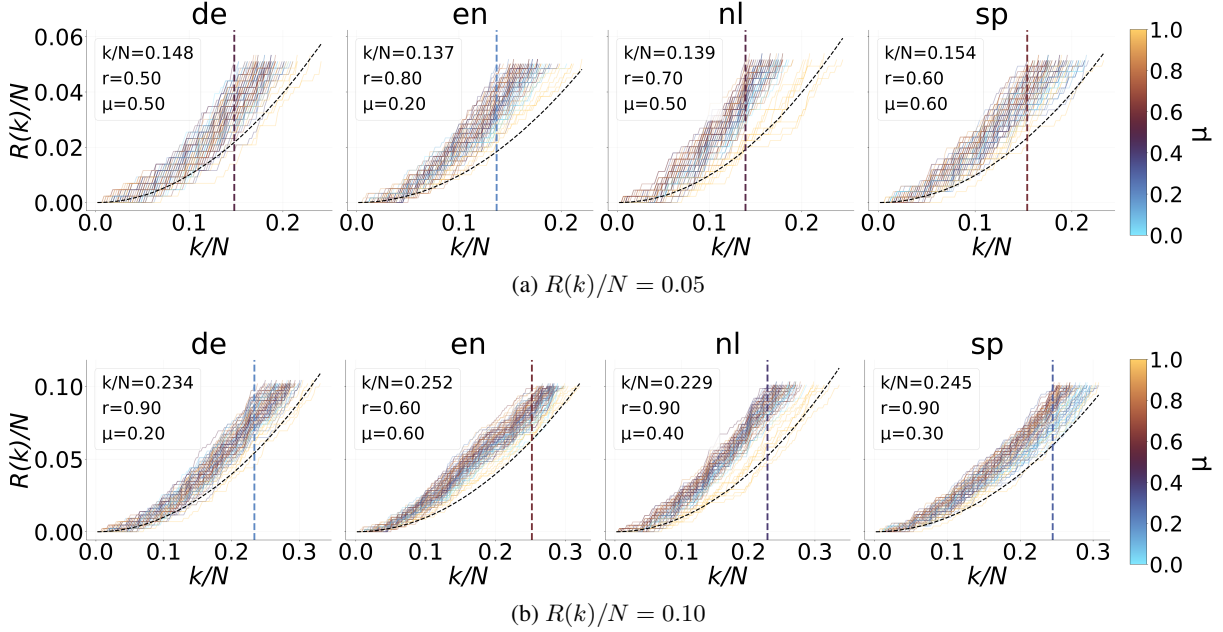

\centering
    \subfloat[$R(k)/N = 0.05$]{\includegraphics[scale=0.14]{fig_recall_0.05.png}}
    \hfill
    \subfloat[$R(k)/N = 0.10$]{\includegraphics[scale=0.14]{fig_recall_0.10.png}}
\caption{\textit{Structural dynamics under spreading activation.} The plots show the ratio of explored nodes $k/N$ required to reach a target $R(k)/N$, where $R(k)$ is the number of empirical words that has to be recovered within the first $k$ positions of the predicted sequence. The vertical dashed line highlights the model requiring the least exploration of the real sequence, with its $r$ and $\mu$ reported in the legend. The black dashed line shows the random acquisition baseline.}
\label{fig:spread_rank}
\end{figure}

\begin{figure}[t!]
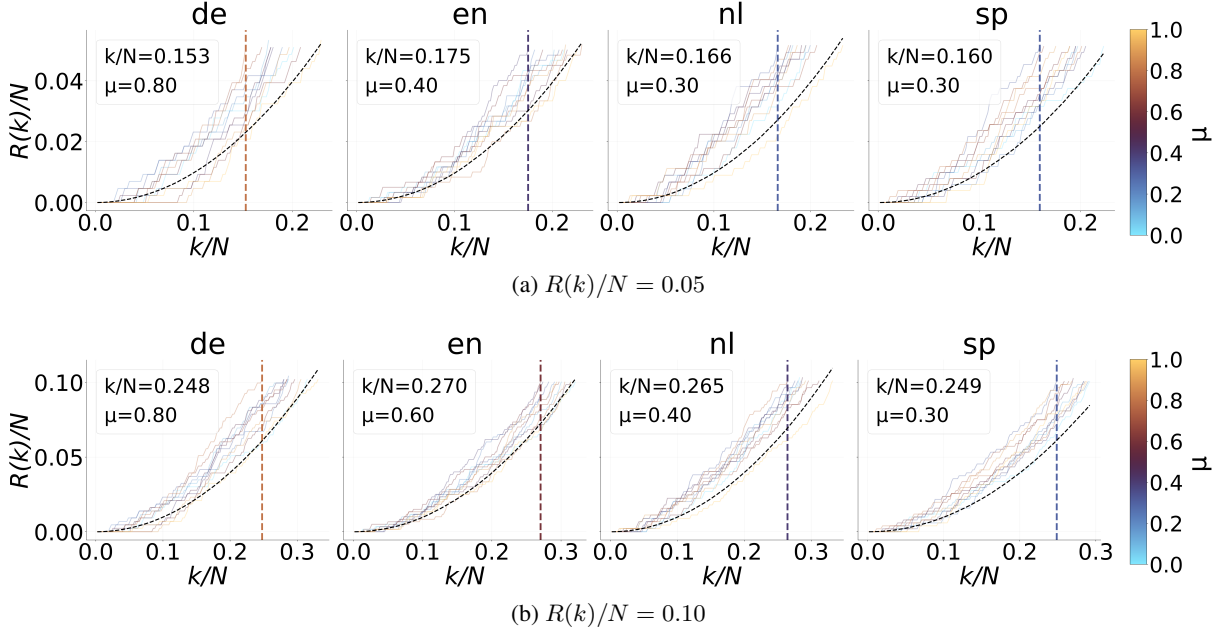

\centering
    \subfloat[$R(k)/N = 0.05$]{\includegraphics[scale=0.14]{fig_recall_geo_0.05.png}}
    \hfill
    \subfloat[$R(k)/N = 0.10$]{\includegraphics[scale=0.14]{fig_recall_geo_0.10.png}}
\caption{\textit{Structural dynamics under shortest paths.} The plots show the ratio of explored nodes $k/N$ required to reach a target $R(k)/N$, where $R(k)$ is the number of empirical words that has to be recovered within the first $k$ positions of the predicted sequence. The vertical dashed line highlights the model requiring the least exploration of the real sequence, with its $\mu$ reported in the legend. The black dashed line shows the random acquisition baseline.}
\label{fig:geo_rank}
\end{figure}

%\FloatBarrier
\subsection{Spreading dynamics outperforms shortest paths in simulating normative sequences}
\label{subsec:recall_sec}

First, we quantify how well the simulated word orderings reproduce the ground-truths.
We measure the fraction of a predicted sequence explored, $k/N$, required to reach a target empirical coverage, $R(k)/N$, where $R(k)$ is the number of empirical words that has to be recovered within the first $k$ positions of the predicted sequence, and $N$ is the total number of words (simplifying the $|N|$ notation for readability).
Conceptually, we address the following question: \textit{How much of the predicted sequence must be explored in order to recover a target fraction of the ground-truth sequence?}

Ideally, recovering a fraction $x$ of the empirical sequence would require exploring exactly the same fraction $x$ on the predicted one, i.e., $k/N = R(k)/N$.
Practically, $k$ exceeds $R(k)$, since some predicted words do not correspond to the empirical ones.
Recovering $R(k)=100$ words may require exploring the $k=150$ or $k=200$ positions of the predicted sequence, with the former $k$ indicating better performance.
Hence, the smaller $k/N$, the better the model.
Moreover, as $R(k)$ increases, any model will eventually reach full coverage.

We consider evaluating portions of $R(k)/N$ in the range $[0.05, 0.85]$ with bins of $0.05$ (17 portions for language, 68 overall), motivated by the hypothesis that different cognitive mechanisms may dominate at different stages of vocabulary development \cite{hills2009longitudinal, storkel2009developmental, citraro2023feature}.
For each $R(k)/N$ portion, we compute the difference between the best performing models with the two structural scores, $\Delta = (k/N)_{\text{spread}} - (k/N)_{\text{shp}}$, with negative values indicating an advantage of spreading activation.
Note that for ``best performing model'' we refer here to the parameter configuration (involving retention $r$ and trade-off $\mu$) that reaches a given recovery level $R(k)/N$ with the smallest exploration cost $k/N$.
Hence, this does not involve parameter fitting or model selection in a supervised learning manner, but only an evaluation with respect to different regimes (the $R(k)/N$ portions). 
We test $H_0:\mathrm{median}(\Delta)=0$ vs $H_1:\mathrm{median}(\Delta)<0$ using a Wilcoxon signed-rank test, leading us to reject $H_0$ ($p<0.01$), and we report Cliff's effect size d $= P(\Delta\ge0)-P(\Delta<0)$ normalized by the number of comparisons $n$ ($d=-0.73$).

Just to provide an intuitive illustration of this evaluation, Figure \ref{fig:spread_rank} and Figure \ref{fig:geo_rank} focus specifically on the first fractions: $R(k)/N=0.05$ and $0.10$.
As an example to read the plots, from the top-left panel of Figure \ref{fig:spread_rank}, recovering $R(k)/N=0.05$ of the dataset for German requires exploring about $k/N=0.148$ of the predicted sequence with the best performing spreading activation model corresponding to $r = 0.5$ and $\mu = 0.5$.
All the other $(r,\mu)$ combinations decrease model performance.
Figure \ref{fig:geo_rank}, instead, shows the results based on shortest paths, and indicates that the best model, i.e., $\mu = 0.8$, requires exploring about $k/N=0.153$ of the predicted sequence.
Being $0.148 < 0.153$, the model encoding the spreading activation score achieves the same level of recovery ($0.05$) with a smaller predicted fraction.

To recap, the statistically significant tendency toward negative differences indicates that spreading activation reproduces the empirical sequences better.
This, in turn, suggests that a dynamics based on shortest paths without an inner temporal scale fails to capture the observed acquisition patterns as effectively as a diffusion process based on explicit temporal propagation throughout the network.

\begin{table}[t!]
\centering
\begin{tabular*}{\tblwidth}{@{}lcccc@{}}
\toprule
 & de & en & nl & sp \\
\midrule
$r$ & -0.35 & -0.50* & -0.56* & -0.58* \\
$\mu$ & 0.49* & -0.22 & 0.03 & 0.25 \\
\bottomrule
\end{tabular*}
\caption{\textit{Structural dynamics under spreading activation.} Spearman's rank correlations between model parameters (respectively, retention $r$ and trade-off $\mu$) and the proportion of explored nodes $k/N$ measured on different $R(k)/N$ vocabulary coverages. A * denotes statistical significance at $p<0.05$.}
\label{tab:corr}
\end{table}

Another potentially interesting observation of the model under spreading activation (see Figure~\ref{fig:spread_rank}) is that the retention of the best performing configurations seems to be relatively high in those two early portions explored, with values lying between $r=0.6$ and $r=0.9$, whereas the trade-off $\mu$ seems to show a more heterogeneous behavior.
Hence, we verify whether this pattern also holds across a wider range of recall levels.
We again consider fractions $R(k)/N$ in the range $[0.05, 0.85]$ with bins of $0.05$.
For each language, we measure the Spearman rank correlation between retention $r$ and $k/N$, and between trade-off $\mu$ and $k/N$.
The correlations are summarized in Table \ref{tab:corr}.
Negative values with respect to retention indicate that higher retention is associated with lower $k/N$, thus better prediction.
This association is statistically significant in English, Dutch, and Spanish.
The result suggests that maintaining high lexical energy when activating words yields better performance than transferring most of the activation to adjacent neighbors.
Instead, the relationship between the trade-off parameter $\mu$ and $k/N$ is less consistent across languages.
Only German shows a statistically significant positive correlation.
However, this indicates that stronger biases toward lexical category exploration, i.e., high $\mu$ values, are associated with worse predictive performance.
The remaining languages highlight non-significant correlations, suggesting that the influence of $\mu$ on recovering normative sequences is limited compared to retention.

\begin{figure}[t!]
\centering
    {\includegraphics[scale=0.43]{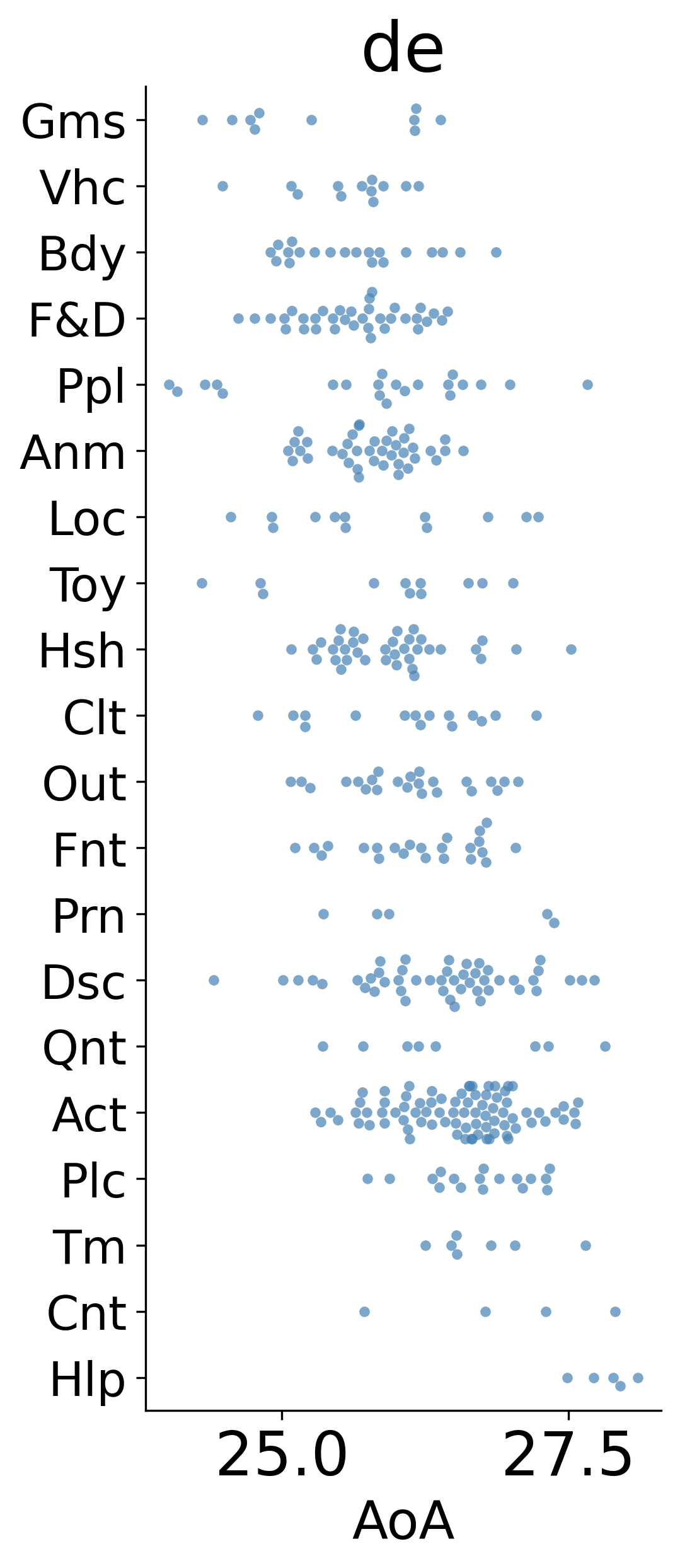}
    \includegraphics[scale=0.43]{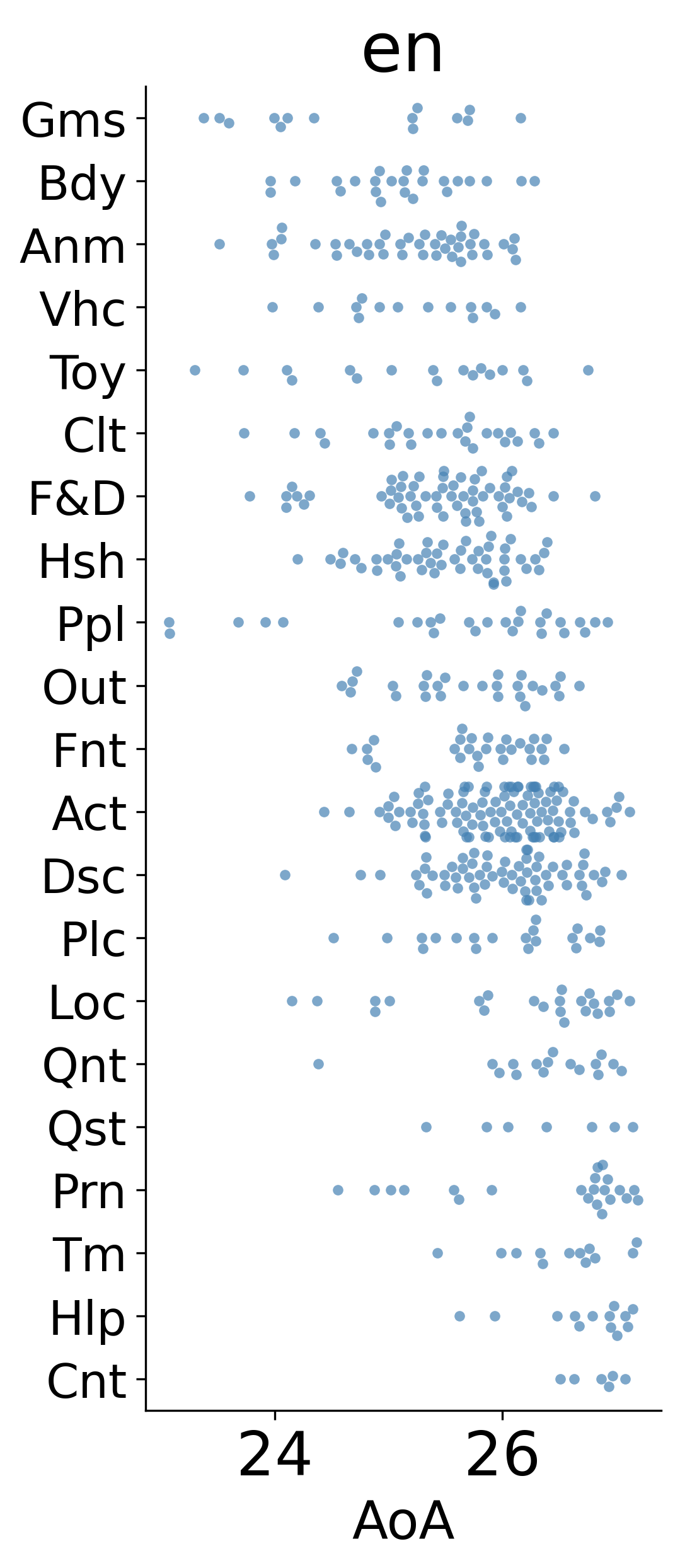}
    \includegraphics[scale=0.43]{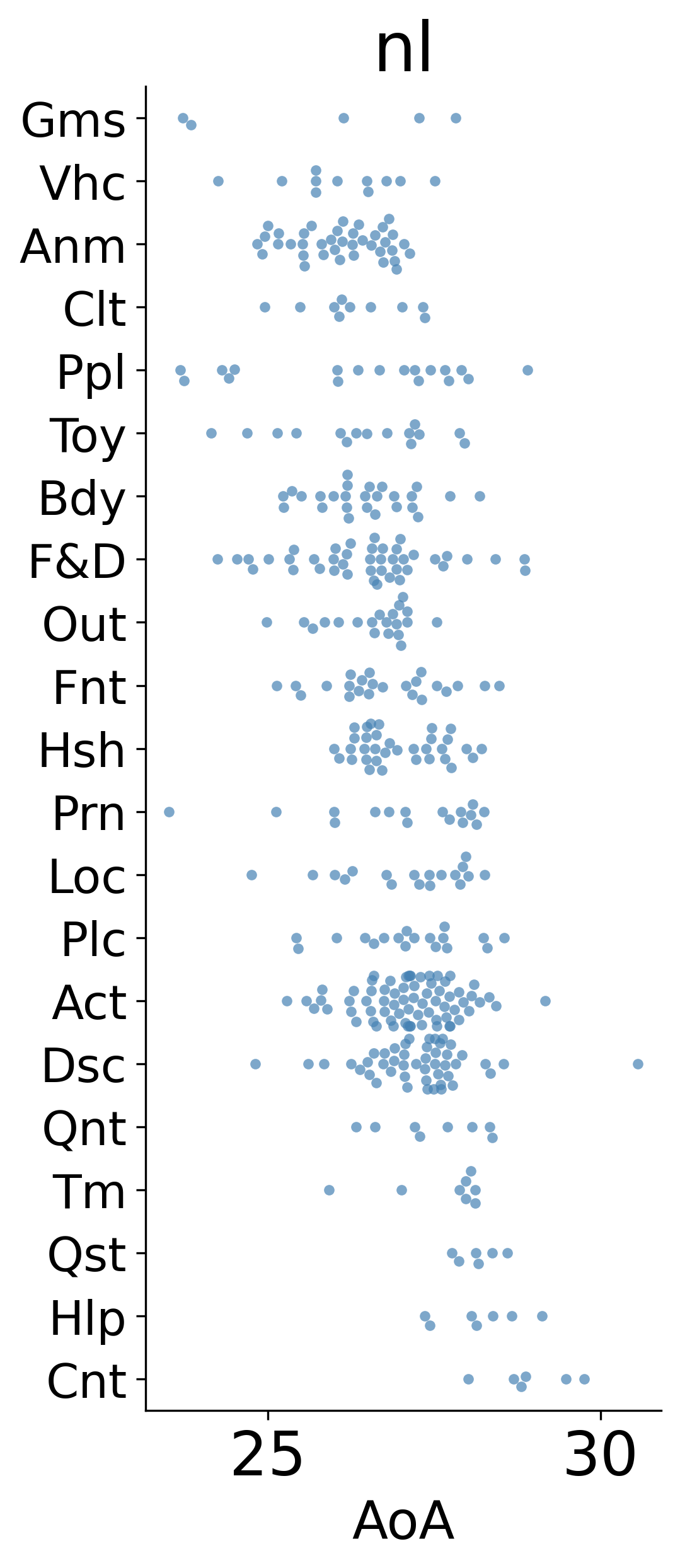}
    \includegraphics[scale=0.43]{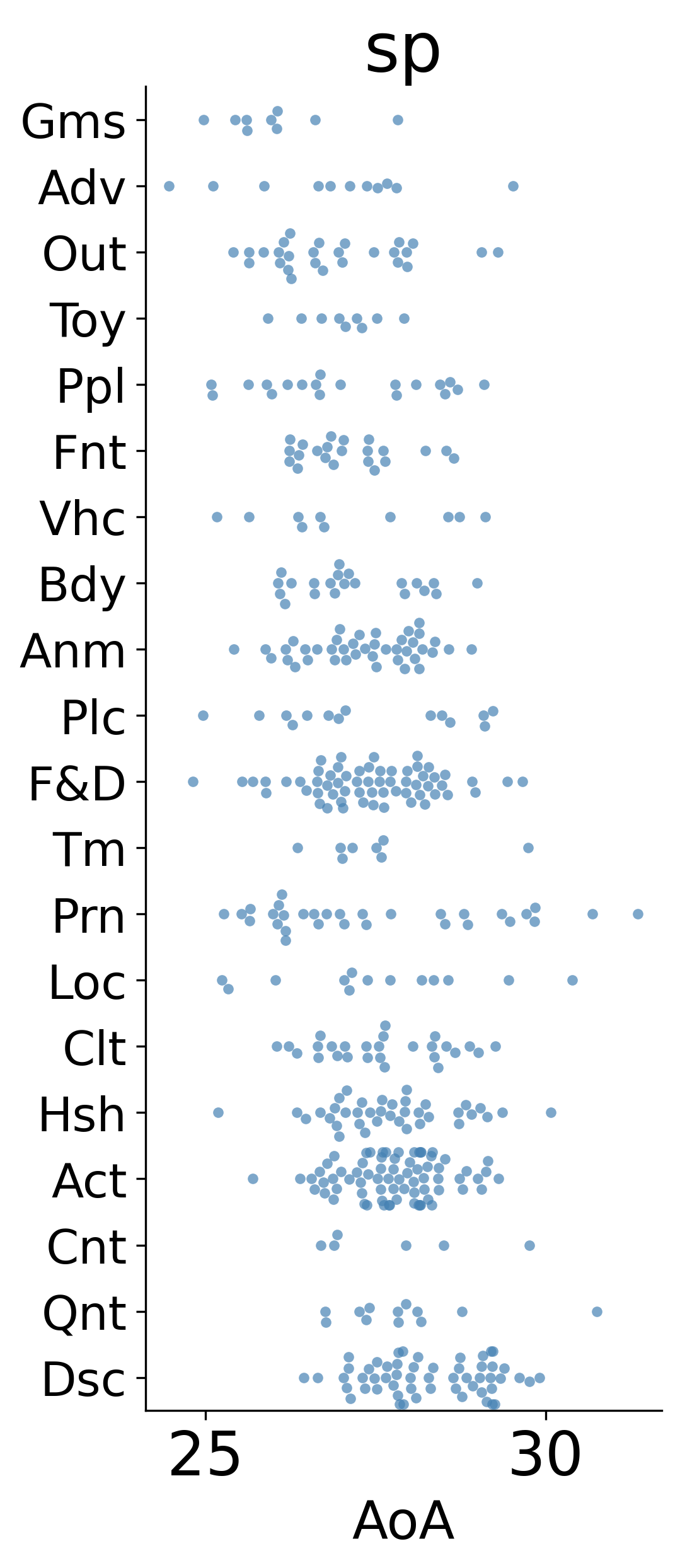}}
\caption{Average normative age of acquisition (in months) per CDI, shown in ascending order.
\textit{Category abbreviations --- Act: Action Words; Adv: Adverbs; Anm: Animals; Bdy: Body Parts; Clt: Clothing; Cnt: Connecting Words; Dsc: Descriptive Words; F\&D: Food \& Drink; Fnt: Furniture \& Rooms; Gms: Games \& Routines; Hlp: Helping Verbs; Hsh: Household; Loc: Locations; Out: Outside; Plc: Places; Ppl: People; Prn: Pronouns; Qnt: Quantifiers; Qst: Question Words; Snd: Sounds; Tm: Time Words; Toy: Toys; Vhc: Vehicles}.}
\label{fig:real_age_and_patterns}
\end{figure}

\begin{figure}[t!]
\centering
    {\includegraphics[scale=0.5]{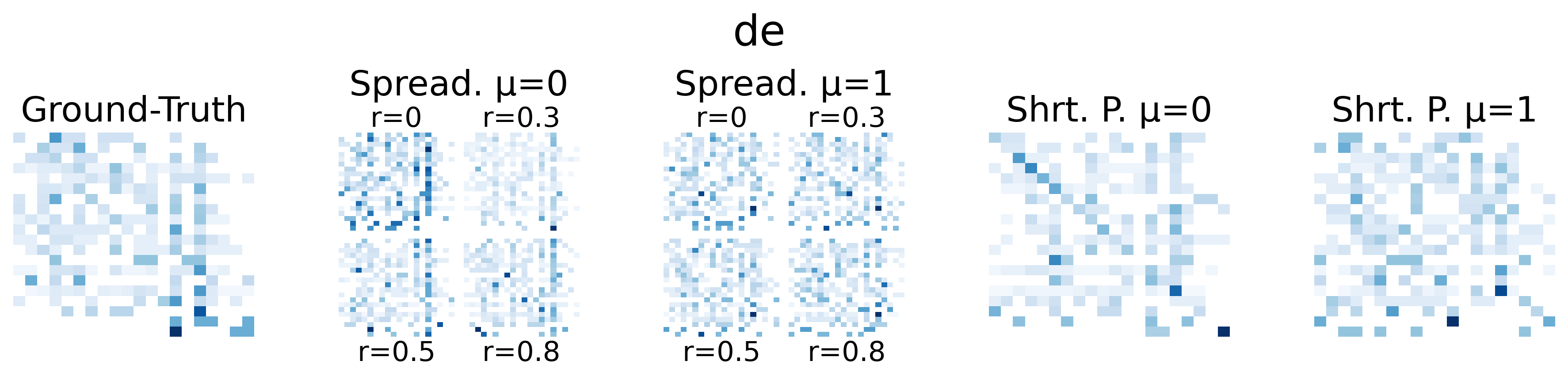}
    \includegraphics[scale=0.5]{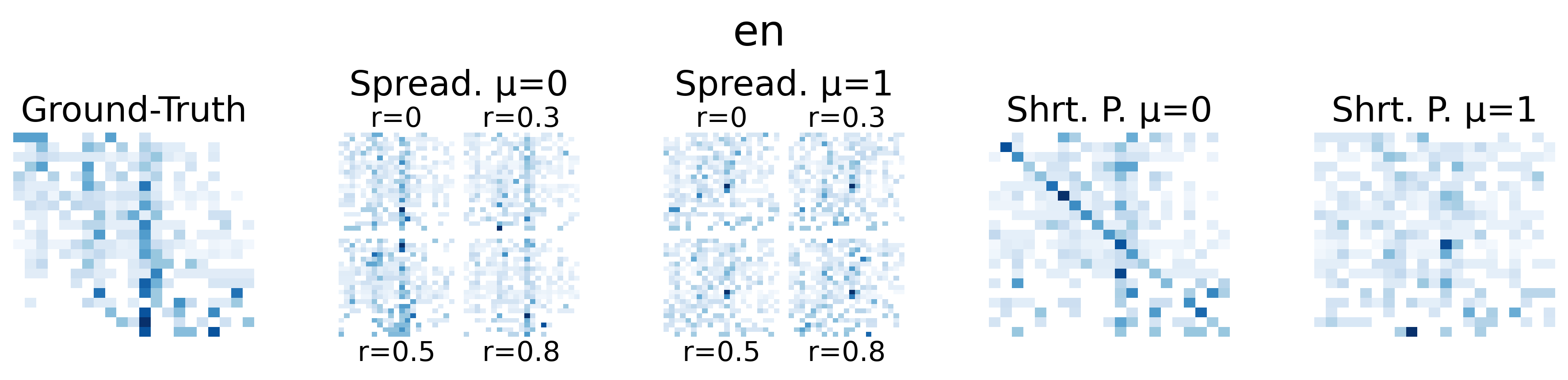}
    \includegraphics[scale=0.5]{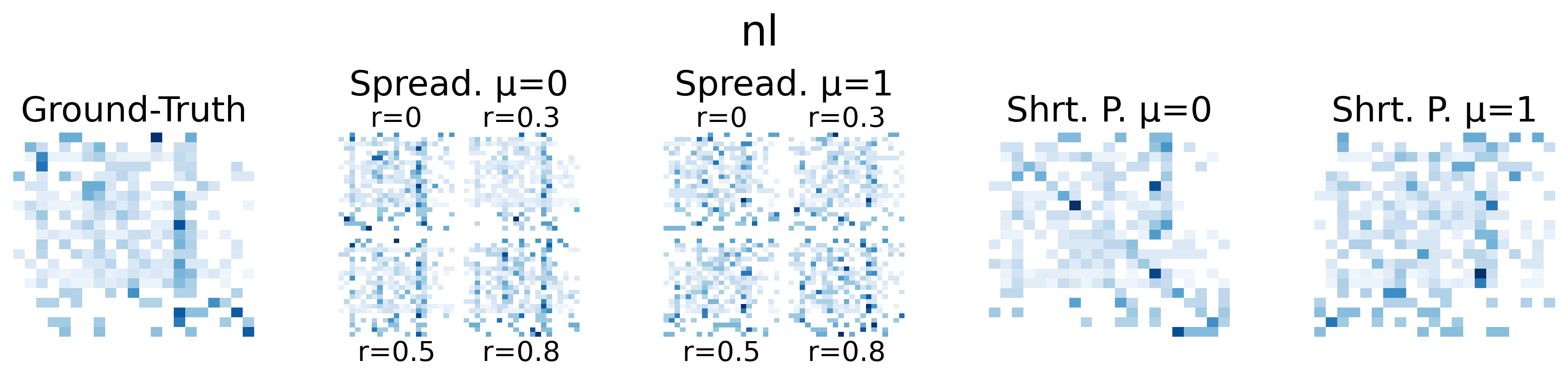}
    \includegraphics[scale=0.5]{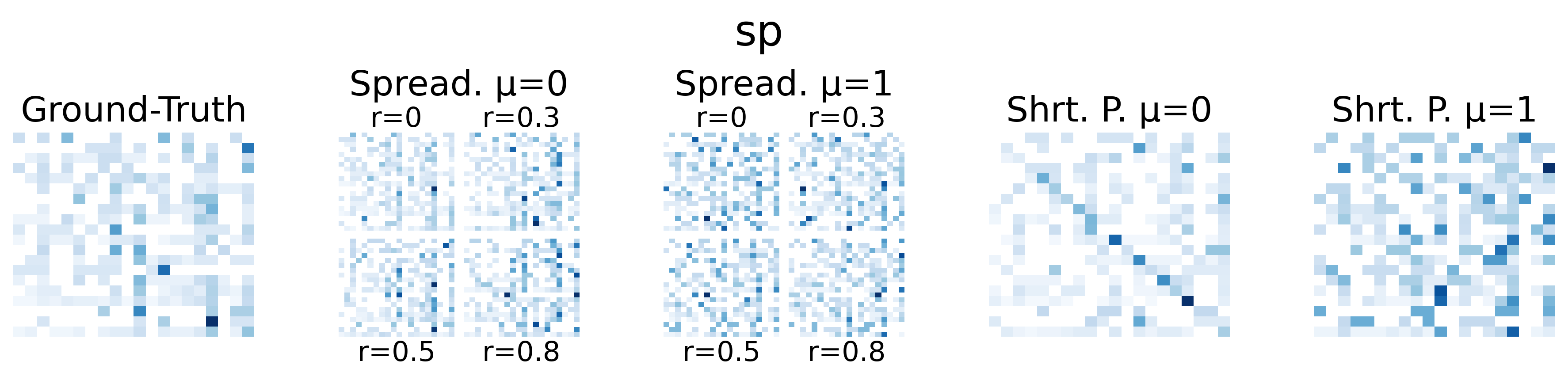}}
\caption{Transition matrices between CDIs, using the same ordering as in Figure \ref{fig:real_age_and_patterns}. Labels are omitted for readability. White cells indicate no transitions, darkness is proportional to probability.}
\label{fig:mix_transition_matrices}
\end{figure}

\begin{figure}[t!]
\centering
    \subfloat[]{\includegraphics[scale=0.32]{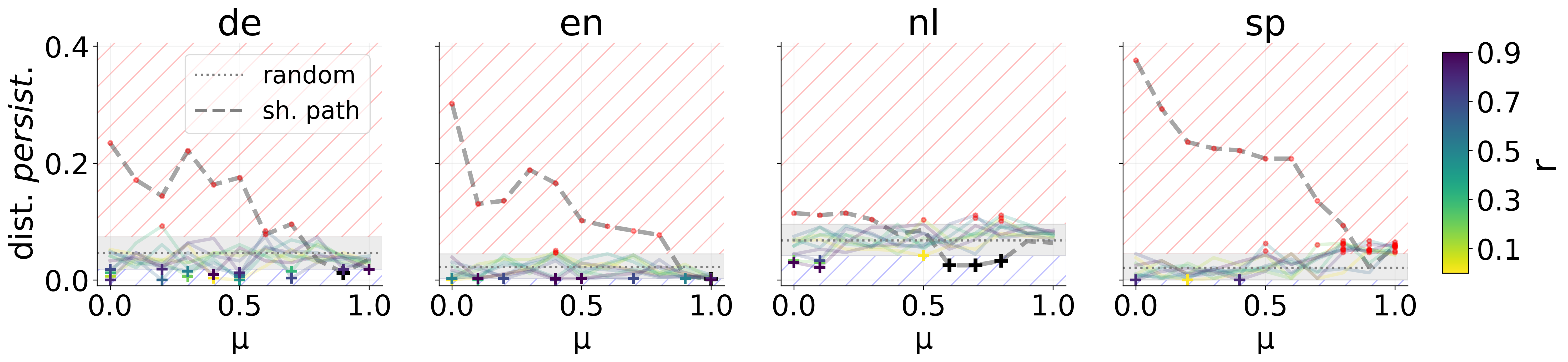}}
    \qquad
    \subfloat[]{\includegraphics[scale=0.32]{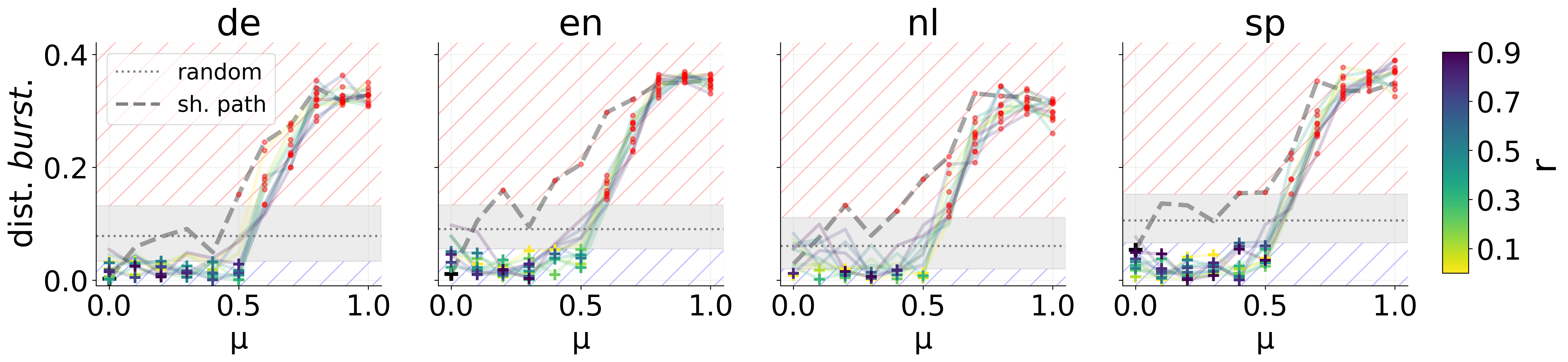}}
\caption{Distances between ground-truth and simulated persistence (a) and burstiness (b) as a function of the parameter $\mu$. Red and blue areas identify statistically significant regions ($p<0.05)$, but only points falling within the blue area indicate model configurations closer to the ground-truth sequences.}
\label{fig:pers_burst}
\end{figure}

%\FloatBarrier
\subsection{Lexical exploration highlighted by CDI transition sequences}
\label{subsec:cdi_res}

We now evaluate the predicted sequences at the level of lexical categories, i.e., the ground-truth CDI forms.
Our objective is to assess whether the dynamics of acquisition with respect to the broader level of CDIs is preserved in the simulated data.
In detail, we examine transitions between consecutive words, focusing on which CDIs tend to follow each other throughout the vocabulary sequence.
To this end, we compare ground-truth and simulated data, assessing the extent to which the latter is similar to the former in terms of i) the average time spent exploiting the same CDI, and ii) burstiness, cf. later.

This analysis stems from the patterns observed in Figure \ref{fig:real_age_and_patterns} and Figure \ref{fig:mix_transition_matrices}.
Across all languages, some of the earliest acquired words belong to the CDIs of people and games, followed by a ``pause'' during which other CDIs are explored before returning to people and games later, see Figure \ref{fig:real_age_and_patterns}.
Similar patterns are also observed for other CDIs.
This suggests non-trivial jumps between categories, which can be characterized by the distribution of inter-times between CDIs, namely the time occurring between the learning of two words belonging to the same CDI, cf. burstiness, later.

Similarly, we are interested in observing whether empirical sequences exhibit exploration or exploitation dynamics.
In Figure \ref{fig:mix_transition_matrices} (ground-truth, leftmost panels), CDI sequences are represented as transition matrices in which each cell indicates the probability of acquiring a word belonging to category $c_j$, given that the previously acquired word belongs to category $c_i$.
Within-CDI transition probabilities are low in ground-truth data, supporting our assumption based on lexical exploration.
Interestingly, by looking at the shortest path baseline, when $\mu=0$ (i.e., when lexical exploration is switched off), we observe a tendency to persist within the same CDI.
This persistency seems not to be captured by the model with the spreading activation score, regardless of retention.
Note that the transition matrices are shown here only for visualization purposes and are not used for the measures introduced in the following.

However, before presenting the measures, Figures \ref{fig:real_age_and_patterns} and \ref{fig:mix_transition_matrices} highlight two other particular aspects.
First, CDI frequencies are heterogeneously distributed; for example, action words are the most frequent CDI category across all languages, while there are few instances of question or connecting words.
Second, since CDIs are ordered by their average age of acquisition, we observe how several categories (such as helping verbs, connecting words, and question words) tend to be acquired later in all languages, and appear to form a relatively cohesive cluster, with transitions among themselves.

To better investigate all these observations, we rely on the following quantities, persistence and burstiness, both measured on the CDI sequences.
Recall that $\mathcal{C} = \{c_1, c_2, \ldots, c_K\}$ denotes the set of $K$ lexical categories or CDIs, $R = (v_1, v_2, \ldots, v_{|N|})$ denotes a (generic) word ordering, thus a CDI sequence $\mathcal{S} = (s_1, s_2, \ldots, s_{|N|})$, with $s_i = c(v_i) \in \mathcal{C}$, is the corresponding sequence of lexical categories obtained by mapping each word onto its CDI.

Persistence $P(\mathcal{S})$ quantifies how long the sequence tends to exploit (i.e., remain within) the same CDI before switching to another CDI.
We identify the maximal subsequences of identical CDIs $s_{max}$ and define persistence as their average length,
$P(\mathcal{S}) = \textit{average length of } s_{max}$.
For example, in $\mathcal{S}=(cat1,cat1,cat1,cat2,cat3,cat3)$, there are three subsequences, of length $3$, $1$, and $2$, respectively, with $P(\mathcal{S})=2$.
This measure ranges from $1$, indicating the highest exploration dynamics, up to $|N|$, ideally.
Concretely, higher values indicate exploitation of CDIs.

Burstiness $B$ is the normalized variation $\omega$, which quantifies the dispersion of inter-times distributions.
An inter-time distribution $\tau$ describes the times that elapse before a target CDI appears again in the learning sequence, thus the dispersion of the distribution is $\omega=\frac{\sigma_{\tau}}{\bar{m}_{\tau}}$, with $\sigma$ indicating the standard deviation, and $\bar{m}$, the mean.
Burstiness, following its original definition \cite{goh2008burstiness}, is defined as follows: $B=\frac{\sigma_{\tau}-\bar{m}_{\tau}}{\sigma_{\tau}+\bar{m}_{\tau}} = \frac{\omega-1}{\omega+1}$.
The measures ranges from -1 to 1, where 1 indicates the bursty CDI sequence ($\sigma_{\tau}$ much higher than $\bar{m}_{\tau}$), -1 the periodic one, and 0 indicates a random sequence in relation to CDIs.
Since inter-time distributions $\tau$ are computed separately for each CDI, we average $B$ across all CDIs, thus obtaining the overall burstiness $B(\mathcal{S})$ of the sequence.

Now we can quantify how close a given sequence is to the ground-truth by the absolute distance between their persistence and burstiness values, e.g., $d(\mathcal{S}_{gt}, \mathcal{S}_{pred}) = \left| P(\mathcal{S}_{gt}) - P(\mathcal{S}_{pred}) \right|$, for persistence.
We compute this distance both between the ground-truth sequence $\mathcal{S}_{gt}$ and each predicted sequence $\mathcal{S}_{pred}$ (with spreading activation and shortest paths as structural scores), and between $\mathcal{S}_{gt}$ and each of the $n=2000$ reshuffled sequences described in Section \ref{subsec:setup}, thus obtaining, in this latter case, an empirical null distribution of distances.
For each language and parameter configuration, we test whether the predicted sequence resembles the ground-truth significantly more than expected under random reshuffling, i.e., we test $H_0$: $d(\mathcal{S}_{gt}, \mathcal{S}_{pred})$ is drawn from the null distribution, against $H_1$: it is smaller than what the null distribution would typically produce.
We compute a one-tailed permutation $p$-value (i.e., distance statistically smaller), and consider $p<0.05$ for significance.
Statistically significant configurations therefore indicate that a predicted sequence matches the empirical persistence and burstiness of CDI transitions more closely than expected under a random relabeling of the same categories.

\begin{figure}[t!]
\centering
    \subfloat[]{\includegraphics[scale=0.32]{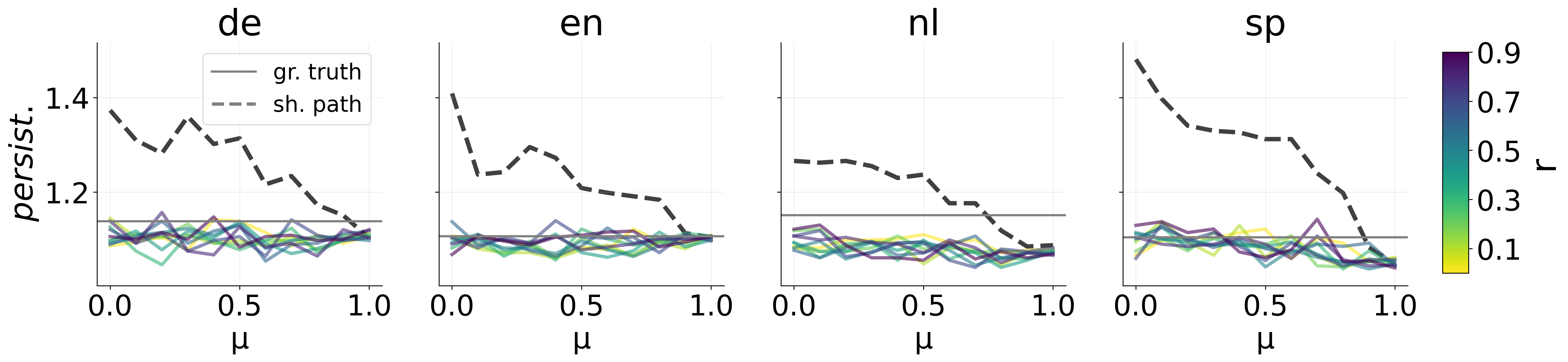}}
    \qquad
    \subfloat[]{\includegraphics[scale=0.32]{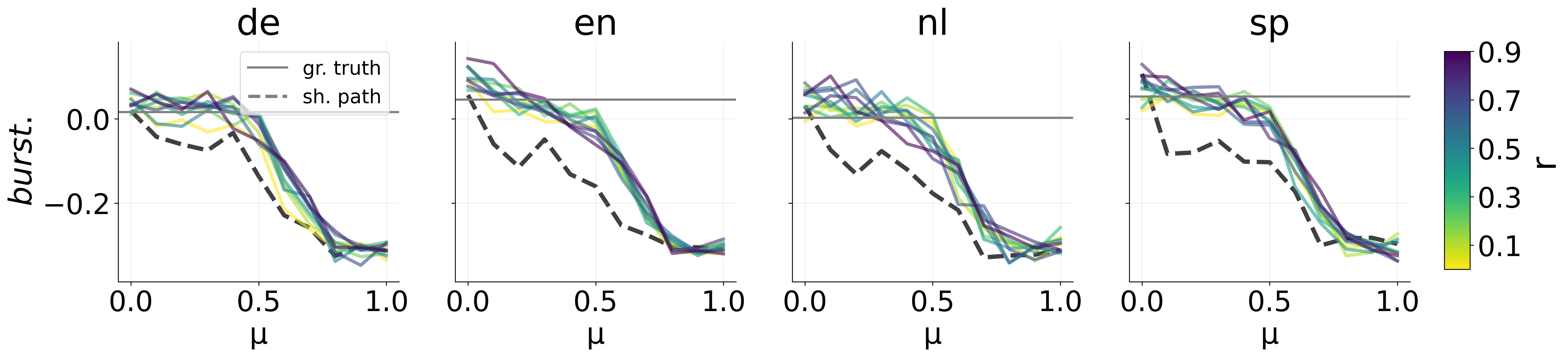}}
\caption{Persistence (c) and burstiness (d) values compared to the ground-truths, horizontal line.}
\label{fig:pers_burst_real}
\end{figure}

Figure \ref{fig:pers_burst} shows the results of the statistical comparison.
In the plots, although only the one-tailed test for greater similarity is used to assess model performance (the blue areas), for clarity we also show (in red) the configurations that are significantly less similar to the ground truth than expected by chance, highlighting the worst performing parameter settings.
This way, we can detect how the model based on shortest paths exhibits the least similar persistence values compared to ground-truths, panel (a).
Dutch is a partial exception, with 3 (out of 11) statistically significant configurations in the direction of greater similarity (the blue area), with $\mu$ ranging from 0.6 to 0.8.
German and English have only one statistically significant configuration each, at $\mu=0.9$ and $\mu=1$, respectively, while Spanish shows no statistically significant configurations.
Instead, spreading activation produces 19 (out of 110) statistically significant configurations for German, 17 for English, and 6 and 3 for Dutch and Spanish, respectively.
These results are associated with Figure \ref{fig:pers_burst_real}, panel (a), where we show the actual persistence values, with the shortest path score overestimating the real persistence, thus indicating exploitation.
The exception of Dutch is also explained by the fact that it is the language with the highest ground-truth persistence among the four languages.

Burstiness results are reported in Figure \ref{fig:pers_burst}, panel (b).
Again, the model leveraging shortest paths exhibits only one (out of 11) statistically significant configuration for German, English, and Spanish, all corresponding to $\mu=0.1$.
Instead, spreading activation highlights 46 (out of 110) statistically significant configurations for both German and English, 26 for Dutch, and 57 for Spanish.
Interestingly, burstiness highlights a clearer pattern than persistence.
None of the spreading activation configurations with $\mu > 0.5$ is statistically significant, suggesting that enforcing exploration dynamics can reproduce the empirical sequences only up to a certain point.
Figure \ref{fig:pers_burst_real}, panel (b), provides further insight to interpret this pattern.
The average burstiness decreases as a function of $\mu$, easily explained by the semantic exploration mechanism $\delta_v^{(t)}$ itself.
By definition, this mechanism pushes the dynamics towards ``empty" CDIs, which translates into an anti-bursty regime characterized by negative values of the measures.

%%--------------------------------------------------------------------
\section{Discussion and Conclusion}\label{sec:disc}
%%--------------------------------------------------------------------

Building on research on spreading activation and network-based lexical acquisition, in this work we have proposed an algorithm for vocabulary development as an iterative search process driven by diffusion dynamics and constraints on lexical categories \cite{collins1975spreading, de2019small, stella2017multiplex, siew2019cognitive, siew2019spreadr, citraro2023feature, citraro2026spreadpy}.
In detail, our model integrates (i) a structural score that models the temporal propagation of lexical activation through a graph-based mental lexicon \cite{collins1975spreading, siew2019cognitive, citraro2026spreadpy}, and (ii) an exploration mechanism reflecting a tendency to expand vocabulary into previously underrepresented semantic domains.
%These two components can be seen as operating in opposition.
%Spreading activation drives the propagation toward structurally neighboring concepts, whereas lexical category exploration can shift this process toward network regions that are distant from the currently activated network neighborhood.
Their interplay is of particular interest, as it might reflect a balance between associative density and inhibitory effects \cite{arias2013s} or skyhook mechanisms \cite{hills2022mind} that have been theoretically proposed but not yet incorporated into computational models of early language acquisition.
It is also consistent with recent studies suggesting that higher-order connectivity may play a more important role than shortest paths in shaping semantic behavior \cite{rubinosimilarity}.

By adopting a cross-linguistic perspective, this framework allows us to analyze early language learning with complex networks beyond studies centered on English \cite{siew2019cognitive, stella2024cognitive, haim2026cognitive}.
Across four languages (German, English, Dutch, and Rioplatense Spanish), our results show that diffusion dynamics outperform both random and shortest path distance baseline models in recovering the empirical acquisition sequences, see Section~\ref{subsec:recall_sec}.
This suggests that structural centrality alone measured on complex mental lexicons is not sufficient to explain early word acquisition sequences, even when combined with a dynamics based on lexical exploration.
Instead, the way activation spreads through a network is more explanatory when placed in a trade-off with the same dynamics of exploration.
Notably, best performing configurations are characterized by high retention levels, which may be linked to neighbor facilitation effects already observed in theoretical models of lexical growth \cite{hills2009longitudinal, hills2010associative} or to priming as a direct effect of spreading activation \cite{collins1975spreading, neely2012semantic}.

At the level of CDI transitions (see Section~\ref{subsec:cdi_res}), our strategy with spreading activation outperforms the baselines in terms of persistence and burstiness under specific parameter regimes.
Transitions at the level of CDIs reveal that persistence times within the same category are, on average, much shorter than those produced by the shortest path baseline (excepting for Dutch), while spreading activation better captures similar dynamics of exploration.
This is further amplified by burstiness, where it is also evident that a strong enforcement of the $\mu$ parameter degrades model performance.
Overall, analyzing transitions between CDIs is a promising path toward uncovering the hidden mechanisms of early language acquisition.

Among the limitations of the present work, two deserve particular attention.
First, high-quality ground-truth data and cross-linguistic comparability may be difficult to obtain.
Although the setting is cross-linguistic, the analysis is limited to four languages, particularly due to the reliance on free association data, that are currently available for only a limited set of languages.
Second, network-based representations of the mental lexicon may have inherent constraints, since they cannot easily capture aspects of early language learning that do not depend on lexical similarity.
Moreover, our model relies on adult lexical networks \cite{de2019small, aeschbach2026small, cabana2024small, de2013better}, which may not accurately reflect the structure of the developing child lexicon \cite{arias2013s}, for which no fully grounded empirical reconstruction is currently available.
That said, in our idea, the role of the network is not to represent what the child already knows, but to structure the space of linguistic candidates from which the next word to be acquired is drawn.
Indeed, this space considers candidate words that are likely to occur in the surrounding speech but have not yet been encoded as stable lexical entries \cite{hills2009longitudinal}.
A closely related limitation to this aspect concerns alternative approaches to network construction, e.g., those based on co-occurrence statistics derived from child-directed speech \cite{stella2017multiplex}.
While exploring co-occurrence layers remains a natural direction for future work, it can introduce non-trivial methodological choices cross-linguistically, such as different values of co-occurrence windows, frequency thresholds, corpus size, etc.
We did not address all these aspects in this work, suspecting that such parameters might vary more across languages than the resources we adopted.

In the future, several directions could extend this study.
A natural improvement would be to incorporate the explicit role of time into the network representation, e.g., by adopting dynamic lexical network models \cite{latapy2018stream} able to account not only for dynamical processes \textit{on} networks (e.g., spreading activation) but also for the evolution \textit{of} the network structure itself.
This would open the possibility of studying how the network topology evolves along with the vocabulary, or also linking the evolution of CDIs to mesoscale-level tracking methods \cite{rossetti2018community}.
Another promising direction is to fully exploit the multilayer nature of lexical representations \cite{stella2017multiplex}, without collapsing all layers into a single aggregated graph, and including co-occurrence layers as well.
This would allow the disentanglement of layer-specific effects \cite{storkel2009developmental}, e.g., by testing hierarchical accounts of lexical development from semantic to phonological organization \cite{angulo2025hierarchical}, as well as the joint influence of layers by enabling the study of superdiffusion, as analyzed in recent work on spreading activation \cite{citraro2026spreadpy}, but not yet applied to early language learning.

%%--------------------------------------------------------------------
\section*{Acknowledgements}

\noindent We would thank Massimo Stella for feedback on the learning algorithm, and Giulio Rossetti for feedback on the results.

\section*{Data and Code Availability}

\noindent All datasets used in this study are publicly available through the original publications cited in Sections 2.1 and 2.2, and we also provide them at the following link repository: \url{https://doi.org/10.5281/zenodo.20700038}.

\noindent The code used to compute the spreading activation structural score is publicly available at: \url{https://github.com/dsalvaz/SpreadPy}.
The remaining can be reproduced from the methodological details provided in Section 2.3.

%\FloatBarrier
%\subsection*{Conflict of interest}
%\noindent The author declares no conflict of interest.

%%--------------------------------------------------------------------

\bibliographystyle{plainnat}
\bibliography{references}

\begin{thebibliography}{45}
\providecommand{\natexlab}[1]{#1}
\providecommand{\url}[1]{\texttt{#1}}
\expandafter\ifx\csname urlstyle\endcsname\relax
  \providecommand{\doi}[1]{doi: #1}\else
  \providecommand{\doi}{doi: \begingroup \urlstyle{rm}\Url}\fi

\bibitem[Aeschbach et~al.(2026)Aeschbach, Mata, L{\~o}o, De~Deyne, and Wulff]{aeschbach2026small}
Samuel Aeschbach, Rui Mata, Kaidi L{\~o}o, Simon De~Deyne, and Dirk~U Wulff.
\newblock The" small world of words" german free-association norms.
\newblock \emph{arXiv preprint arXiv:2604.19620}, 2026.

\bibitem[Angulo-Chavira et~al.(2025)Angulo-Chavira, Castell{\'o}n-Flores, and Arias-Trejo]{angulo2025hierarchical}
Armando~Quetzalc{\'o}atl Angulo-Chavira, Alejandra~Mitzi Castell{\'o}n-Flores, and Natalia Arias-Trejo.
\newblock Hierarchical prediction in toddlers: Semantic and phonological development.
\newblock \emph{Journal of Memory and Language}, 145:\penalty0 104688, 2025.

\bibitem[Arias-Trejo and Plunkett(2013)]{arias2013s}
Natalia Arias-Trejo and Kim Plunkett.
\newblock What’s in a link: Associative and taxonomic priming effects in the infant lexicon.
\newblock \emph{Cognition}, 128\penalty0 (2):\penalty0 214--227, 2013.

\bibitem[Beckage et~al.(2011)Beckage, Smith, and Hills]{beckage2011small}
Nicole Beckage, Linda Smith, and Thomas Hills.
\newblock Small worlds and semantic network growth in typical and late talkers.
\newblock \emph{PloS one}, 6\penalty0 (5):\penalty0 e19348, 2011.

\bibitem[Beckage et~al.(2012)Beckage, Steyvers, and Butts]{beckage2012route}
Nicole Beckage, Mark Steyvers, and Carter Butts.
\newblock Route choice in individuals—semantic network navigation.
\newblock In \emph{Proceedings of the Annual Meeting of the Cognitive Science Society}, volume~34, 2012.

\bibitem[Beckage and Colunga(2016)]{beckage2016language}
Nicole~M Beckage and Eliana Colunga.
\newblock Language networks as models of cognition: Understanding cognition through language.
\newblock \emph{Towards a theoretical framework for analyzing complex linguistic networks}, pages 3--28, 2016.

\bibitem[Beckage and Colunga(2019)]{beckage2019network}
Nicole~M Beckage and Eliana Colunga.
\newblock Network growth modeling to capture individual lexical learning.
\newblock \emph{Complexity}, 2019\penalty0 (1):\penalty0 7690869, 2019.

\bibitem[Bloom(2002)]{bloom2002}
Paul Bloom.
\newblock \emph{How Children Learn the Meanings of Words}.
\newblock MIT Press, Cambridge, MA, 2002.

\bibitem[Bond and Foster(2013)]{bond2013linking}
Francis Bond and Ryan Foster.
\newblock Linking and extending an open multilingual wordnet.
\newblock In \emph{Proceedings of the 51st Annual Meeting of the Association for Computational Linguistics (Volume 1: Long Papers)}, pages 1352--1362, 2013.

\bibitem[Braginsky et~al.(2019)Braginsky, Yurovsky, Marchman, and Frank]{braginsky2019consistency}
Mika Braginsky, Daniel Yurovsky, Virginia~A Marchman, and Michael~C Frank.
\newblock Consistency and variability in children’s word learning across languages.
\newblock \emph{Open Mind}, 3:\penalty0 52--67, 2019.

\bibitem[Cabana et~al.(2024)Cabana, Zugarramurdi, Valle-Lisboa, and De~Deyne]{cabana2024small}
{\'A}lvaro Cabana, Camila Zugarramurdi, Juan~C Valle-Lisboa, and Simon De~Deyne.
\newblock The" small world of words" free association norms for rioplatense spanish.
\newblock \emph{Behavior Research Methods}, 56\penalty0 (2):\penalty0 968--985, 2024.

\bibitem[Chern et~al.(2025)Chern, Castro, and Siew]{chern2025evidence}
Jazton Chern, Nichol Castro, and Cynthia~SQ Siew.
\newblock Evidence of community structure in phonological networks of multiple languages.
\newblock \emph{Canadian Journal of Experimental Psychology/Revue canadienne de psychologie exp{\'e}rimentale}, 79\penalty0 (1):\penalty0 4, 2025.

\bibitem[Citraro et~al.(2023)Citraro, Vitevitch, Stella, and Rossetti]{citraro2023feature}
Salvatore Citraro, Michael~S Vitevitch, Massimo Stella, and Giulio Rossetti.
\newblock Feature-rich multiplex lexical networks reveal mental strategies of early language learning.
\newblock \emph{Scientific Reports}, 13\penalty0 (1):\penalty0 1474, 2023.

\bibitem[Citraro et~al.(2026)Citraro, Haim, Carini, Siew, Rossetti, and Stella]{citraro2026spreadpy}
Salvatore Citraro, Edith Haim, Alessandra Carini, Cynthia~SQ Siew, Giulio Rossetti, and Massimo Stella.
\newblock Spreadpy: A python tool for modelling spreading activation and superdiffusion in cognitive multiplex networks.
\newblock \emph{Computers in Human Behavior Reports}, page 100940, 2026.

\bibitem[Collins and Loftus(1975)]{collins1975spreading}
Allan~M Collins and Elizabeth~F Loftus.
\newblock A spreading-activation theory of semantic processing.
\newblock \emph{Psychological review}, 82\penalty0 (6):\penalty0 407, 1975.

\bibitem[De~Deyne et~al.(2013)De~Deyne, Navarro, and Storms]{de2013better}
Simon De~Deyne, Daniel~J Navarro, and Gert Storms.
\newblock Better explanations of lexical and semantic cognition using networks derived from continued rather than single-word associations.
\newblock \emph{Behavior research methods}, 45\penalty0 (2):\penalty0 480--498, 2013.

\bibitem[De~Deyne et~al.(2019)De~Deyne, Navarro, Perfors, Brysbaert, and Storms]{de2019small}
Simon De~Deyne, Danielle~J Navarro, Amy Perfors, Marc Brysbaert, and Gert Storms.
\newblock The “small world of words” english word association norms for over 12,000 cue words.
\newblock \emph{Behavior research methods}, 51\penalty0 (3):\penalty0 987--1006, 2019.

\bibitem[De~Saussure(2017)]{de2017course}
Ferdinand De~Saussure.
\newblock Course in general linguistics.
\newblock \emph{Literary theory: An anthology}, pages 137--152, 2017.

\bibitem[Fenson et~al.(2007)]{fenson2007macarthur}
Larry Fenson et~al.
\newblock Macarthur-bates communicative development inventories.
\newblock 2007.

\bibitem[Ferrer-i Cancho and Vitevitch(2018)]{ferrer2018origins}
Ramon Ferrer-i Cancho and Michael~S Vitevitch.
\newblock The origins of zipf's meaning-frequency law.
\newblock \emph{Journal of the Association for Information Science and Technology}, 69\penalty0 (11):\penalty0 1369--1379, 2018.

\bibitem[Firth(1957)]{firth1957synopsis}
John Firth.
\newblock A synopsis of linguistic theory, 1930-1955.
\newblock \emph{Studies in linguistic analysis}, pages 10--32, 1957.

\bibitem[Goh and Barab{\'a}si(2008)]{goh2008burstiness}
K-I Goh and A-L Barab{\'a}si.
\newblock Burstiness and memory in complex systems.
\newblock \emph{Europhysics Letters}, 81\penalty0 (4):\penalty0 48002, 2008.

\bibitem[Haim and Stella(2026)]{haim2026cognitive}
Edith Haim and Massimo Stella.
\newblock Cognitive networks for knowledge modeling: A gentle introduction for data-and cognitive scientists.
\newblock \emph{Wiley Interdisciplinary Reviews: Cognitive Science}, 17\penalty0 (2):\penalty0 e70026, 2026.

\bibitem[Hills and Kenett(2022)]{hills2022mind}
Thomas~T Hills and Yoed~N Kenett.
\newblock Is the mind a network? maps, vehicles, and skyhooks in cognitive network science.
\newblock \emph{Topics in Cognitive Science}, 14\penalty0 (1):\penalty0 189--208, 2022.

\bibitem[Hills et~al.(2009)Hills, Maouene, Maouene, Sheya, and Smith]{hills2009longitudinal}
Thomas~T Hills, Mounir Maouene, Josita Maouene, Adam Sheya, and Linda Smith.
\newblock Longitudinal analysis of early semantic networks: Preferential attachment or preferential acquisition?
\newblock \emph{Psychological science}, 20\penalty0 (6):\penalty0 729--739, 2009.

\bibitem[Hills et~al.(2010)Hills, Maouene, Riordan, and Smith]{hills2010associative}
Thomas~T Hills, Josita Maouene, Brian Riordan, and Linda~B Smith.
\newblock The associative structure of language: Contextual diversity in early word learning.
\newblock \emph{Journal of memory and language}, 63\penalty0 (3):\penalty0 259--273, 2010.

\bibitem[Hills et~al.(2012)Hills, Jones, and Todd]{hills2012optimal}
Thomas~T Hills, Michael~N Jones, and Peter~M Todd.
\newblock Optimal foraging in semantic memory.
\newblock \emph{Psychological review}, 119\penalty0 (2):\penalty0 431, 2012.

\bibitem[Hills et~al.(2015)Hills, Todd, and Jones]{hills2015foraging}
Thomas~T Hills, Peter~M Todd, and Michael~N Jones.
\newblock Foraging in semantic fields: How we search through memory.
\newblock \emph{Topics in cognitive science}, 7\penalty0 (3):\penalty0 513--534, 2015.

\bibitem[Latapy et~al.(2018)Latapy, Viard, and Magnien]{latapy2018stream}
Matthieu Latapy, Tiphaine Viard, and Cl{\'e}mence Magnien.
\newblock Stream graphs and link streams for the modeling of interactions over time.
\newblock \emph{Social Network Analysis and Mining}, 8\penalty0 (1):\penalty0 61, 2018.

\bibitem[Lenci et~al.(2022)Lenci, Sahlgren, Jeuniaux, Cuba~Gyllensten, and Miliani]{lenci2022comparative}
Alessandro Lenci, Magnus Sahlgren, Patrick Jeuniaux, Amaru Cuba~Gyllensten, and Martina Miliani.
\newblock A comparative evaluation and analysis of three generations of distributional semantic models.
\newblock \emph{Language resources and evaluation}, 56\penalty0 (4):\penalty0 1269--1313, 2022.

\bibitem[Neely(2012)]{neely2012semantic}
James~H Neely.
\newblock Semantic priming effects in visual word recognition: A selective review of current findings and theories.
\newblock \emph{Basic processes in reading}, pages 264--336, 2012.

\bibitem[Rossetti and Cazabet(2018)]{rossetti2018community}
Giulio Rossetti and R{\'e}my Cazabet.
\newblock Community discovery in dynamic networks: a survey.
\newblock \emph{ACM computing surveys (CSUR)}, 51\penalty0 (2):\penalty0 1--37, 2018.

\bibitem[Rubino and Piazza()]{rubinosimilarity}
Valerio Rubino and Manuela Piazza.
\newblock Similarity and generalization as discounted integration over higher-order paths in semantic networks.

\bibitem[Siegel and Bond(2021)]{siegel2021odenet}
Melanie Siegel and Francis Bond.
\newblock Odenet: Compiling a germanwordnet from other resources.
\newblock In \emph{Proceedings of the 11th global wordnet conference}, pages 192--198, 2021.

\bibitem[Siew(2019)]{siew2019spreadr}
Cynthia~SQ Siew.
\newblock spreadr: An r package to simulate spreading activation in a network.
\newblock \emph{Behavior Research Methods}, 51\penalty0 (2):\penalty0 910--929, 2019.

\bibitem[Siew et~al.(2019)Siew, Wulff, Beckage, and Kenett]{siew2019cognitive}
Cynthia~SQ Siew, Dirk~U Wulff, Nicole~M Beckage, and Yoed~N Kenett.
\newblock Cognitive network science: A review of research on cognition through the lens of network representations, processes, and dynamics.
\newblock \emph{Complexity}, 2019\penalty0 (1):\penalty0 2108423, 2019.

\bibitem[Stella et~al.(2017)Stella, Beckage, and Brede]{stella2017multiplex}
Massimo Stella, Nicole~M Beckage, and Markus Brede.
\newblock Multiplex lexical networks reveal patterns in early word acquisition in children.
\newblock \emph{Scientific reports}, 7\penalty0 (1):\penalty0 46730, 2017.

\bibitem[Stella et~al.(2024)Stella, Citraro, Rossetti, Marinazzo, Kenett, and Vitevitch]{stella2024cognitive}
Massimo Stella, Salvatore Citraro, Giulio Rossetti, Daniele Marinazzo, Yoed~N Kenett, and Michael~S Vitevitch.
\newblock Cognitive modelling of concepts in the mental lexicon with multilayer networks: Insights, advancements, and future challenges.
\newblock \emph{Psychonomic Bulletin \& Review}, 31\penalty0 (5):\penalty0 1981--2004, 2024.

\bibitem[Steyvers and Tenenbaum(2005)]{steyvers2005large}
Mark Steyvers and Joshua~B Tenenbaum.
\newblock The large-scale structure of semantic networks: Statistical analyses and a model of semantic growth.
\newblock \emph{Cognitive science}, 29\penalty0 (1):\penalty0 41--78, 2005.

\bibitem[Storkel(2009)]{storkel2009developmental}
Holly~L Storkel.
\newblock Developmental differences in the effects of phonological, lexical and semantic variables on word learning by infants.
\newblock \emph{Journal of child language}, 36\penalty0 (2):\penalty0 291--321, 2009.

\bibitem[Utsumi(2015)]{utsumi2015complex}
Akira Utsumi.
\newblock A complex network approach to distributional semantic models.
\newblock \emph{PloS one}, 10\penalty0 (8):\penalty0 e0136277, 2015.

\bibitem[Vitevitch(2008)]{vitevitch2008can}
Michael~S Vitevitch.
\newblock What can graph theory tell us about word learning and lexical retrieval?
\newblock \emph{Journal of speech, language, and hearing research}, 51\penalty0 (2):\penalty0 408--422, 2008.

\bibitem[Vitevitch and Luce(2016)]{vitevitch2016phonological}
Michael~S Vitevitch and Paul~A Luce.
\newblock Phonological neighborhood effects in spoken word perception and production.
\newblock \emph{Annual review of linguistics}, 2:\penalty0 75--94, 2016.

\bibitem[Weizman and Snow(2001)]{weizman2001lexical}
Zehava~Oz Weizman and Catherine~E Snow.
\newblock Lexical output as related to children's vocabulary acquisition: Effects of sophisticated exposure and support for meaning.
\newblock \emph{Developmental psychology}, 37\penalty0 (2):\penalty0 265, 2001.

\bibitem[Wittgenstein(1953)]{wittgenstein1953}
Ludwig Wittgenstein.
\newblock \emph{Philosophical Investigations}.
\newblock Blackwell Publishers, Oxford, 1953.

\end{thebibliography}

\end{document}